\documentclass{article}

\usepackage{microtype}
\usepackage{graphicx}
\usepackage{subfigure}
\usepackage{booktabs} 

\usepackage{hyperref}

\usepackage[accepted]{arxiv_icml}

\usepackage{tikz}
\usepackage{amsmath}
\usepackage{epsfig}
\usepackage{graphicx}
\usepackage{amsmath}
\usepackage{amssymb}
\usepackage{comment}
\usepackage{multirow}
\usepackage{tabularx}
\usepackage{tabu}
\usepackage{array}
\usepackage{xspace}
\usepackage{pgfplots}
\pgfplotsset{compat=1.13}
\usepackage{sansmath}
\newcommand{\steg}{\textsc{steganogan}\xspace}
\newcommand{\bpp}{bpp\xspace}
\newcommand{\bestbpp}{4.4\xspace}
\newcommand{\githuburl}{\url{https://github.com/DAI-Lab/SteganoGAN}}

\icmltitlerunning{SteganoGAN}

\makeatletter
\renewcommand{\printAffiliationsAndNotice}[1]{%
{\let\thefootnote\relax\footnotetext{\hspace*{-\footnotesep}\ifdefined\isaccepted #1\fi%
\ifdefined\icmlcorrespondingauthor@text
Correspondence to: \icmlcorrespondingauthor@text.
\else
{\bf AUTHORERR: Missing \textbackslash{}icmlcorrespondingauthor.}
\fi

\ \\
\Notice@String
}
}
}
\makeatother

\begin{document}

\twocolumn[
\icmltitle{SteganoGAN: High Capacity Image Steganography with GANs}



\icmlsetsymbol{equal}{*}

\begin{icmlauthorlist}
\textbf{\large Kevin A. Zhang,\textsuperscript{1} Alfredo Cuesta-Infante,\textsuperscript{2} Lei Xu,\textsuperscript{1} Kalyan Veeramachaneni\textsuperscript{1}}
\par\vspace{0.5em}
\textsuperscript{1} MIT, Cambridge, MA - 02139, USA\\
\texttt{kevz,leix,kalyanv@mit.edu}\\
\textsuperscript{2} Univ. Rey Juan Carlos, Spain\\
\texttt{alfredo.cuesta@urjc.es}
\end{icmlauthorlist}


\icmlcorrespondingauthor{Kevin A. Zhang}{kevz@mit.edu}

\icmlkeywords{Machine Learning, Steganography, ICML}

\vskip 0.3in
]



\printAffiliationsAndNotice{}  

\begin{figure*}[ht!]
\begin{center}
\includegraphics[width=0.195\linewidth]{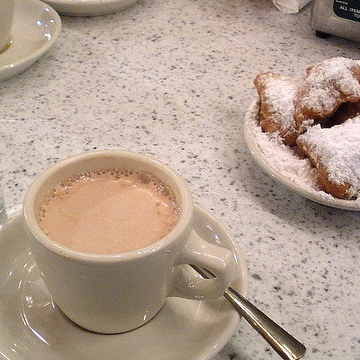}
\includegraphics[width=0.195\linewidth]{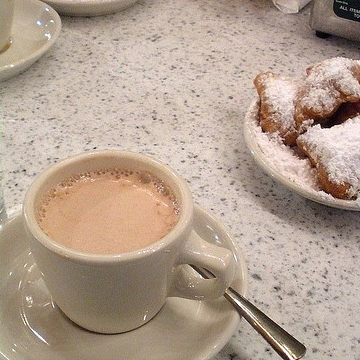}
\includegraphics[width=0.195\linewidth]{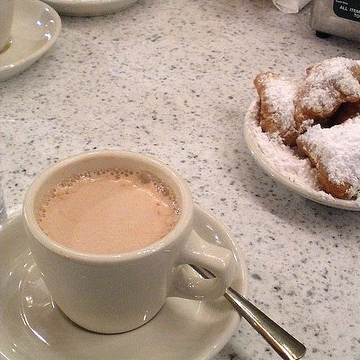}
\includegraphics[width=0.195\linewidth]{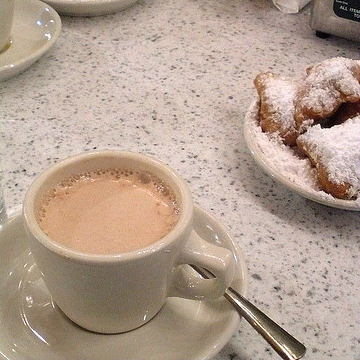}
\includegraphics[width=0.195\linewidth]{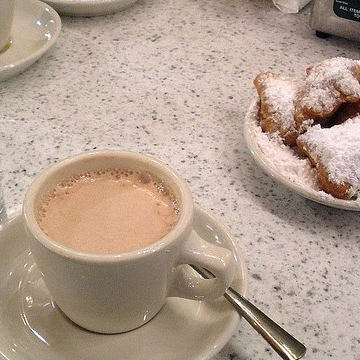}
\caption{A randomly selected cover image (left) and the corresponding steganographic images generated by \steg at approximately 1, 2, 3, and 4 bits per pixel.}
\label{fig:introduction_examples}
\end{center}
\end{figure*}

\begin{abstract}
Image steganography is a procedure for hiding messages inside pictures. While other techniques such as cryptography aim to prevent adversaries from reading the secret message, steganography aims to hide the presence of the message itself. In this paper, we propose a novel technique for hiding arbitrary binary data in images using generative adversarial networks which allow us to optimize the perceptual quality of the images produced by our model. We show that our approach achieves state-of-the-art payloads of \bestbpp bits per pixel, evades detection by steganalysis tools, and is effective on images from multiple datasets. To enable fair comparisons, we have released an open source library that is available online at:
\githuburl.
\end{abstract}

\section{Introduction}
The goal of image steganography is to hide a secret message inside an image. In a typical scenario, the sender hides a secret message inside a cover image and transmits it to the receiver, who recovers the message. Even if the image is intercepted, no one besides the sender and receiver should be able to detect the presence of a message.

Traditional approaches to image steganography are only effective up to a relative payload of around 0.4 bits per pixel \cite{Pevny}. Beyond that point, they tend to introduce artifacts that can be easily detected by automated steganalysis tools and, in extreme cases, by the human eye. With the advent of deep learning in the past decade, a new class of image steganography approaches is emerging \cite{Hayes,Baluja,Zhu}. These approaches use neural networks as either a component in a traditional algorithm (e.g. using deep learning to identify spatial locations suitable for embedding data), or as an end-to-end solution, which takes in a cover image and a secret message and combines them into a steganographic image. 

 These attempts have proved that deep learning can be used for practical end-to-end image steganography, and have achieved embedding rates competitive with those accomplished through traditional techniques \cite{Pevny}. However, they are also more limited than their traditional counterparts: they often impose special constraints on the size of the cover image (for example, \cite{Hayes} requires the cover images to be 32 x 32); they attempt to embed images inside images and not arbitrary messages or bit vectors; and finally, they do not explore the limits of how much information can be hidden successfully. We provide the reader a detailed analysis of these methods in Section~\ref{related}.

To address these limitations, we propose \steg, a novel end-to-end model for image steganography that builds on recent advances in deep learning. We use dense connections which mitigate the vanishing gradient problem and have been shown to improve performance \cite{densenet}. In addition, we use multiple loss functions within an adversarial training framework to optimize our encoder, decoder, and critic networks simultaneously. We find that our approach successfully embeds arbitrary data into cover images drawn from a variety of natural scenes and achieves state-of-the-art embedding rates of \bestbpp bits per pixel while evading standard detection tools. Figure~\ref{fig:introduction_examples} presents some example images that demonstrate the effectiveness of \steg. The left-most figure is the original cover image without any secret messages. The next four figures contain approximately 1, 2, 3, and 4 bits per pixel worth of secret data, respectively, without producing any visible artifacts.

\textbf{Our contributions through this paper are}:
\vspace{-2mm}
\begin{itemize}
\vspace{-2mm}
    \item[--] We present a novel approach that uses adversarial training to solve the steganography task and achieves a relative payload of \bestbpp bits per pixel which is 10x higher than competing deep learning-based approaches with similar peak signal to noise ratios.
    \item[--] We propose a new metric for evaluating the \textit{capacity} of deep learning-based steganography algorithms, which enables comparisons against traditional approaches.
    \item[--] We evaluate our approach by measuring its ability to evade traditional steganalysis tools which are designed to detect whether an image is steganographic or not. Even when we encode $>4$ bits per pixel into the image, most traditional steganalysis tools still only achieve a detection auROC of $<0.6$.
    \item[--] We also evaluate our approach by measuring its ability to evade deep learning-based steganalysis tools. We train a state-of-the-art model for automatic steganalysis proposed by \cite{deep_steganalysis} on samples generated by our model. If we require our model to produce steganographic images such that the detection rate is at most 0.8 auROC, we find that our model can still hide up to 2 bits per pixel.
    \item[--] We are releasing a fully-maintained open-source library called \steg\footnote{\githuburl}, including datasets and pre-trained models, which will be used to evaluate deep learning based steganography techniques.
\end{itemize}

The rest of the paper is organized as follows. Section~\ref{motivates} briefly describes our motivation for building a better image steganography system. Section~\ref{steganogan} presents \steg and describes our model architecture. Section~\ref{metrics} describes our metrics for evaluating model performance. Section~\ref{results} contains our experiments for several variants of our model. Section~\ref{detection} explores the effectiveness of our model at avoiding detection by automated steganalysis tools. Section~\ref{related} details related work in the generation of steganographic images.

\section{Motivation}\label{motivates}
There are several reasons to use steganography instead of (or in addition to) cryptography when communicating a secret message between two actors. First, the information contained in a cryptogram is accessible to anyone who has the private key, which poses a challenge in countries where private key disclosure is required by law. Furthermore, the very existence of a cryptogram reveals the presence of a message, which can invite attackers. These problems with plain cryptography exist in security, intelligence services, and a variety of other disciplines \cite{s12130-003-1026-4}.

For many of these fields, steganography offers a promising alternative. For example, in medicine, steganography can be used to hide private patient information in images such as X-rays or MRIs \cite{CBMS.2004.1311702} as well as biometric data  \cite{s11042-017-5308-3}. In the media sphere, steganography can be used to embed copyright data \cite{j.aeue.2014.11.004} and allow content access control systems to store and distribute digital works over the Internet \cite{Kawaguchi_2007}. In each of these situations, it is important to embed as much information as possible, and for that information to be both \textit{undetectable} and \textit{lossless} to ensure the data can be recovered by the recipient. Most work in the area of steganography, including the methods described in this paper, targets these two goals. We propose a new class of models for image steganography that achieves both these goals.

\section{SteganoGAN} \label{steganogan}
In this section, we introduce our notation, present the model architecture, and describe the training process. At a high level, steganography requires just two operations: \textit{encoding} and \textit{decoding}. The \textit{encoding} operation takes a cover image and a binary message, and creates a steganographic image. The \textit{decoding} operation takes the steganographic image and recovers the binary message.

\subsection{Notation}
\newcommand{\Conv}{\mathtt{Conv}}
\newcommand{\Cat}{\mathtt{Cat}}
\newcommand{\FConv}{\mathtt{FConv}}
\newcommand{\Mean}{\mathtt{Mean}}
We have $C$ and $S$ as the cover image and the steganographic image respectively, both of which are RGB color images and have the same resolution $W \times H$; let $M\in\{0,1\}^{D\times W\times H}$ be the binary message that is to be hidden in $C$. Note that $D$ is the upper-bound on the relative payload; the actual relative payload is the number of bits that can reliably decoded which is given by $(1-2p)D$, where $p\in[0,1]$ is the error rate. The actual relative payload is discussed in more detail in Section~\ref{metrics}.

The cover image $C$ is sampled from the probability distribution of all natural images $\mathbb{P}_{C}$. The steganographic image $S$ is then generated by a learned encoder $\mathcal{E}(C, M)$. The secret message $\hat{M}$ is then extracted by a learned decoder $\mathcal{D}(S)$. The optimization task, given a fixed message distribution, is to train the encoder $\mathcal{E}$ and the decoder $\mathcal{D}$ to  minimize (1) the decoding error rate $p$ and (2) the distance between natural and steganographic image distributions $dis(\mathbb{P}_{C}, \mathbb{P}_{S})$. Therefore, to optimize the encoder and the decoder, we also need to train a critic network $\mathcal{C}(\cdot)$ to estimate $dis(\mathbb{P}_{C}, \mathbb{P}_{S})$.

Let $X\in\mathbb{R}^{D\times W \times H}$ and $ Y\in\mathbb{R}^{D'\times W \times H}$ be two tensors of the same width and height but potentially different depth, $D$ and $D'$; then, let
$\Cat:(X,Y)\rightarrow \Phi\in\mathbb{R}^{(D+D')\times W \times H}$ be the concatenation of the two tensors along the depth axis.

Let $\Conv_{D \rightarrow D'}:X\in\mathbb{R}^{D\times W \times H}\rightarrow \Phi\in\mathbb{R}^{D'\times W \times H}$ be a convolutional block that maps an input tensor $X$ into a feature map $\Phi$ of the same width and height but potentially different depth. This convolutional block consists of a convolutional layer with kernel size 3, stride 1 and padding `same', followed by a leaky ReLU activation function and batch normalization. The activation function and batch normalization operations are omitted if the convolutional block is the last block in the network.

Let $\Mean:X\in\mathbb{R}^{D \times W \times H}\rightarrow \mathbb{R}^D$ represent the adaptive mean spatial pooling operation which computes the average of the $W\times H$ values in each feature map of tensor X.

\subsection{Architecture}

\begin{figure*}[t]
    \centering
    \begin{tabular}{c}
        \includegraphics[width=0.85\linewidth]{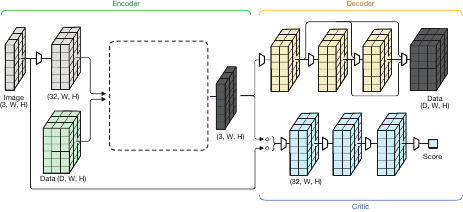} \\
        (a) \\
    \end{tabular}
    \begin{tabular}{ccc}
        \includegraphics[width=.3\linewidth]{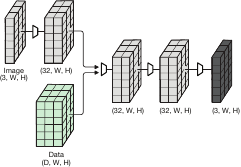} & 
        \includegraphics[width=.3\linewidth]{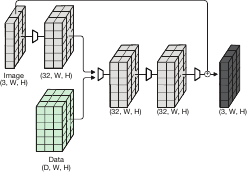}   & 
        \includegraphics[width=.3\linewidth]{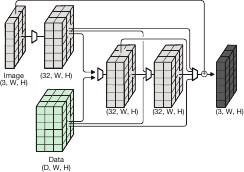} \\
        (b) & (c) & (d) \\
    \end{tabular}
    \caption{(a) The model architecture with the Encoder, Decoder, and Critic. The blank rectangle representing the Encoder can be any of the following: (b) Basic encoder, (c) Residual encoder and (d) Dense encoder. The trapezoids represent convolutional blocks, two or more arrows merging represent concatenation operations, and the curly bracket represents a batching operation.}
    \label{fig:architecture}
\end{figure*}

In this paper, we present \steg, a generative adversarial network for hiding an arbitrary bit vector in a cover image. Our proposed architecture, shown in Figure~\ref{fig:architecture}, consists of three modules: (1) an Encoder that takes a cover image and a data tensor, or message, and produces a steganographic image (Section~\ref{encoder}); (2) a Decoder that takes the steganographic image and attempts to recover the data tensor (Section~\ref{decoder}), and (3) a Critic that evaluates the quality of the cover and steganographic images (Section~\ref{critic}).

\subsubsection{Encoder} \label{encoder}
The encoder network takes a cover image $C$ and a message $M\in\{0,1\}^{D\times W\times H}$. Hence $M$ is a binary data tensor of shape $D \times W \times H$ where $D$ is the number of bits that we will attempt to hide in each pixel of the cover image. 

We explore three variants of the encoder architecture with different connectivity patterns. All the variants start by applying the following two operations:

\begin{enumerate}
    \item Processing the cover image $C$ with a convolutional block to obtain the tensor $a$ given by
    \begin{equation}\label{eq:tensor_a}
        a = \Conv_{3 \rightarrow 32}(C)
    \end{equation}
    \item Concatenating the message $M$ to $a$ and then processing the result with a convolutional block to obtain the tensor $b$:
    \begin{equation}\label{eq:tensor_b}
        b = \Conv_{32+D \rightarrow 32}(\Cat(a,M))
    \end{equation}
\end{enumerate}

\noindent\textbf{Basic:~ }
We sequentially apply two convolutional blocks to tensor $b$ and generate the steganographic image as shown in Figure \ref{fig:architecture}b. Formally:
\begin{equation}\label{eq:formal_encoder_B}
     \mathcal{E}_b(C,M) = \Conv_{32 \rightarrow 3}(\Conv_{32 \rightarrow 32}(b)),
\end{equation}
This approach is similar to that in \cite{Baluja} as the steganographic image is simply the output of the last convolutional block.

\noindent \textbf{Residual}: 
The use of residual connections has been shown to improve model stability and convergence \cite{resnet} so we hypothesize that its use will improve the quality of the steganographic image. To this end we modify the basic encoder by adding the cover image $C$ to its output so that the encoder learns to produce a residual image as shown in Figure~\ref{fig:architecture}c. Formally,
\begin{equation}\label{eq:formal_encoder_C}
     \mathcal{E}_r(C,M) = C + \mathcal{E}_b(C,M),
\end{equation}

\noindent \textbf{Dense}: In the dense variant, we introduce additional connections between the convolutional blocks so that the feature maps generated by the earlier blocks are concatenated to the feature maps generated by later blocks as shown in Figure \ref{fig:architecture}d. This connectivity pattern is inspired by the DenseNet \cite{densenet} architecture which has been shown to encourage feature reuse and mitigate the vanishing gradient problem. Therefore, we hypothesize that the use of dense connections will improve the embedding rate. It can be formally expressed as follows
\begin{equation}\label{eq:formal_encoder_D}
\left\lbrace
\begin{array}{l}
        c = \Conv_{64+D \rightarrow 32}(\Cat(a, b, M)) \\
        d = \Conv_{96+D \rightarrow 3}(\Cat(a, b, c, M)) \\
        \mathcal{E}_d(C,M) = C + d
\end{array}
\right.
\end{equation}

Finally, the output of each variant is a steganographic image $S = \mathcal{E}_{\{b,r,d\}}(C,M)$ that has the same resolution and depth than the cover image $C$.

\subsubsection{Decoder} \label{decoder}
The decoder network takes the steganographic image $S$ produced by the encoder. Formally it can be expressed as:
\begin{equation}\label{eq:formal_decoder}
\left\lbrace
\begin{array}{l}
        a = \Conv_{3 \rightarrow 32}(S) \\
        b = \Conv_{32 \rightarrow 32}(a) \\
        c = \Conv_{64 \rightarrow 32}(\Cat(a, b)) \\
        \mathcal{D}(S) = \Conv_{96 \rightarrow D}(\Cat(a, b, c)) \\
\end{array}
\right.
\end{equation}
The decoder produces $\hat{M} = \mathcal{D}_d(S)$; in other words it attempts to recover the data tensor $M$.

\subsubsection{Critic} \label{critic}
To provide feedback on the performance of our encoder and generate more realistic images, we introduce an adversarial Critic. The critic network consists of three convolutional blocks followed by a convolutional layer with one output channel. To produce the scalar score, we apply adaptive mean pooling to the output of the convolutional layer.

\begin{equation}\label{eq:formal_critic}
\left\lbrace
\begin{array}{l}
        a = \Conv_{32 \rightarrow 32}(\Conv_{32 \rightarrow 32}(\Conv_{3 \rightarrow 32}(S))) \\ 
        \mathcal{C}(S) = \Mean(\Conv_{32 \rightarrow 1}((a))
\end{array}
\right.
\end{equation}

\subsection{Training}
We iteratively optimize the encoder-decoder network and the critic network. To optimize the encoder-decoder network, we jointly optimize three losses: (1) the decoding accuracy using the cross-entropy loss
\begin{equation}
    \mathcal{L}_d = \mathbb{E}_{X\sim \mathbb{P}_{C}} \text{CrossEntropy}(\mathcal{D}(\mathcal{E}(X, M)), M)
\end{equation}
(2) the similarity between steganographic image and the cover image using mean square error
\begin{equation}
    \mathcal{L}_s = \mathbb{E}_{X\sim \mathbb{P}_{C}} \frac{1}{3\times W\times H}||X-\mathcal{E}(X, M)||_2^2
\end{equation}
(3) and the realness of the steganographic image using the critic network 
\begin{equation}
    \mathcal{L}_r = \mathbb{E}_{X\sim \mathbb{P}_{C}} \mathcal{C}(\mathcal{E}(X, M))
\end{equation}
The training objective is to
\begin{equation}
    \text{minimize } \mathcal{L}_d + \mathcal{L}_s + \mathcal{L}_r.
\end{equation} 
To train the critic network, we minimize the Wasserstein loss 
\begin{align}
    \mathcal{L}_c=& \mathbb{E}_{X\sim \mathbb{P}_{C}} \mathcal{C}(X) \nonumber\\
    & -\mathbb{E}_{X\sim \mathbb{P}_{C}} \mathcal{C}(\mathcal{E}(X, M))
\end{align}

During every iteration, we match each cover image $C$ with a data tensor $M$, which consists of a randomly generated sequence of $D\times W\times H$ bits sampled from a Bernoulli distribution $M\sim Ber(0.5)$. In addition, we apply standard data augmentation procedures including horizontal flipping and random cropping to cover image $C$ in our pre-processing pipeline. We use the Adam optimizer with learning rate $1e^{-4}$, clip our gradient norm to $0.25$, clip the critic weights to $[-0.1, 0.1]$, and train for $32$ epochs.

\begin{figure*}
\begin{center}
\includegraphics[width=0.245\linewidth]{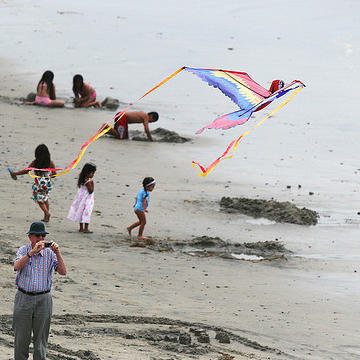}
\includegraphics[width=0.245\linewidth]{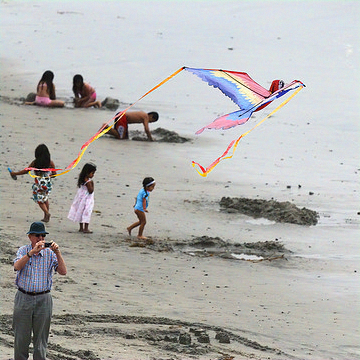}
\includegraphics[width=0.245\linewidth]{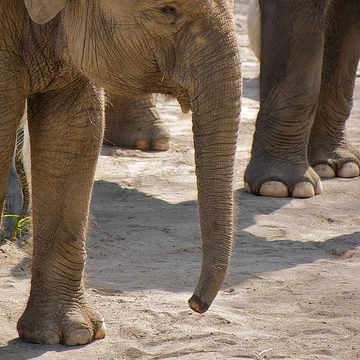}
\includegraphics[width=0.245\linewidth]{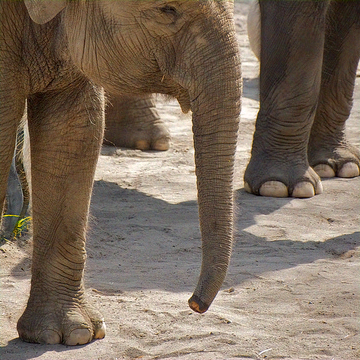}
\includegraphics[width=0.245\linewidth]{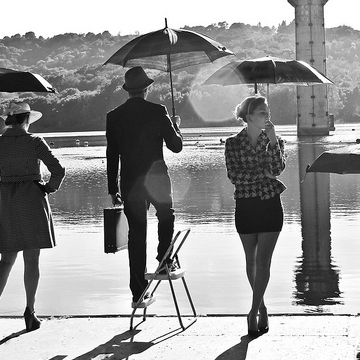}
\includegraphics[width=0.245\linewidth]{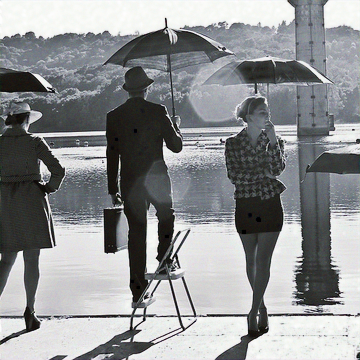}
\includegraphics[width=0.245\linewidth]{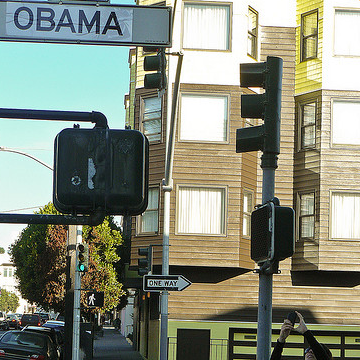}
\includegraphics[width=0.245\linewidth]{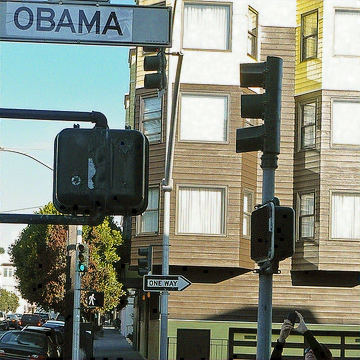}
\caption{Randomly selected pairs of cover (left) and steganographic (right) images from the COCO dataset which embeds random binary data at the maximum payload of \bestbpp bits-per-pixel.}
\label{fig:examples}
\end{center}
\end{figure*}

\section{Evaluation Metrics} \label{metrics} 
Steganography algorithms are evaluated along three axes: the amount of data that can be hidden in an image, a.k.a \textit{capacity}, the similarity between the cover and steganography image, a.k.a \textit{distortion}, and the ability to avoid detection by steganalysis tools, a.k.a \textit{secrecy}. This section describes some metrics for evaluating the performance of our model along these axes.

\noindent \textbf{Reed Solomon Bits Per Pixel}:
Measuring the effective number of bits that can be conveyed per pixel is non-trivial in our setup since the ability to recover a hidden bit is heavily dependent on the model and the cover image, as well as the message itself.

To model this situation, suppose that a given model incorrectly decodes a bit with probability $p$. It is tempting to just multiply the number of bits in the data tensor by the accuracy $1-p$ and report that value as the relative payload. Unfortunately, that value is actually meaningless -- it allows you to estimate the number of bits that have been correctly decoded, but does not provide a mechanism for recovering from errors or even identifying which bits are correct.

Therefore, to get an accurate estimate of the relative payload of our technique, we turn to Reed-Solomon codes. Reed-Solomon error-correcting codes are a subset of linear block codes which offer the following guarantee: Given a message of length $k$, the code can generate a message of length $n$ where $n \geq k$ such that it can recover from $\frac{n-k}{2}$ errors \cite{Reed}. This implies that given a steganography algorithm which, on average, returns an incorrect bit with probability $p$, we would want the number of incorrect bits to be less than or equal to the number of bits we can correct:

\begin{equation}
    \label{eq:ratio}
    p\cdot n \leq \frac{n-k}{2}
\end{equation}

The ratio $k/n$ represents the average number of bits of "real" data we can transmit for each bit of "message" data; then, from (\ref{eq:ratio}), it follows that the ratio is less than or equal to $1-2p$. As a result, we can measure the relative payload of our steganographic technique by multiplying the number of bits we attempt to hide in each pixel by the ratio to obtain the "real" number of bits that is transmitted and recovered.

We refer to this metric as Reed-Solomon bits-per-pixel (RS-BPP), and note that it can be directly compared against traditional steganographic techniques since it represents the average number of bits that can be reliably transmitted in an image divided by the size of the image.

\noindent \textbf{Peak Signal to Noise Ratio}: In addition to measuring the relative payload, we also need to measure the quality of the steganographic image. One widely-used metric for measuring image quality is the peak signal-to-noise ratio (PSNR). This metric is designed to measure image distortions and has been shown to be correlated with mean opinion scores produced by human experts \cite{ssim}.

Given two images $X$ and $Y$ of size $(W,H)$ and a scaling factor $\mathtt{sc}$ which represents the maximum possible difference in the numerical representation of each pixel\footnote{For example, if the images are represented as floating point numbers in $[-1.0, 1.0]$, then $\mathtt{sc}=2.0$ since the maximum difference between two pixels is achieved when one is $1.0$ and the other is $-1.0$.}, the PSNR is defined as a function of the mean squared error (MSE):

\begin{align}
    \mathrm{MSE} &= \frac{1}{W H} \sum_{i=1}^W \sum_{j=1}^H (X_{i,j}-Y_{i,j})^2, \\
    \mathrm{PSNR} &= 20 \cdot \log_{10}(\mathtt{sc}) - 10 \cdot \log_{10}(\mathrm{MSE})
\end{align}

Although PSNR is widely used to evaluate the distortion produced by steganography algorithms, \cite{stego_psnr} suggests that it may not be ideal for comparisons across different types of steganography algorithms. Therefore, we introduce another metric to help us evaluate image quality: the structural similarity index.

\noindent \textbf{Structural Similarity Index}: In our experiments, we also report the structural similarity index (SSIM) between the cover image and the steganographic image. SSIM is widely used in the broadcast industry to measure image and video quality \cite{ssim}. Given two images $X$ and $Y$, the SSIM can be computed using the means, $\mu_X$ and $\mu_Y$, variances, $\sigma_X^2$ and $\sigma_Y^2$ , and covariance $\sigma_{XY}^2$ of the images as shown below:

\begin{equation}
    \mathrm{SSIM} = \frac{(2\mu_X\mu_Y+k_1R) (2\sigma_{XY}+k_2R)}{(\mu_X^2 + \mu_Y^2 + k_1R) (\sigma_X^2+\sigma_Y^2+k_2R)}
\end{equation}

The default configuration for SSIM uses $k_1=0.01$ and $k_2=0.03$ and returns values in the range $[-1.0, 1.0]$ where $1.0$ indicates the images are identical.

\section{Results and Analysis}\label{results}
We use the Div2k \cite{div2k} and COCO \cite{mscoco} datasets to train and evaluate our model. We experiment with each of the three model variants discussed in Section~\ref{steganogan} and train them with 6 different data depths $D \in \{1, 2, ..., 6\}$. The data depth $D$ represents the ``target" bits per pixel so the randomly generated data tensor has shape $D$ x $W$ x $H$.

We use the default train/test split proposed by the creators of the Div2K and COCO data sets in our experiments, and we report the average \texttt{RS-BPP}, PSNR, and SSIM on the test set in Table~\ref{table:metrics}. Our models are trained on GeForce GTX 1080 GPUs. The wall clock time per epoch is approximately 10 minutes for Div2K and 2 hours for COCO.

After training our model, we compute the expected accuracy on a held-out test set and adjust it using the Reed-Solomon coding scheme discussed in Section~\ref{metrics} to produce our bits-per-pixel metric, shown in Table~\ref{table:metrics} under \texttt{RS-BPP}. We publicly released the pre-trained models for all the experiments shown in this table on AWS S3\footnote{\url{http://steganogan.s3.amazonaws.com/}}.

The results from our experiments are shown in Table \ref{table:metrics} -- each of the metrics is computed on a held-out test set of images that is not shown to the model during training. Note that there is an unavoidable tradeoff between the relative payload and the image quality measures; assuming we are already on the Pareto frontier, an increased relative payload would inevitably result in a decreased similarity.

\begin{table*}
\fontsize{9.4}{12}\selectfont
\centering

\begin{tabular}{|r|c|ccc|ccc|ccc|ccc|}
\hline
\multirow{2}{*}{\textbf{Dataset}} & \multirow{2}{*}{\textbf{D}} & \multicolumn{3}{c|}{\textbf{Accuracy}}              & \multicolumn{3}{c|}{\textbf{RS-BPP}}                            & \multicolumn{3}{c|}{\textbf{PSNR}}                  & \multicolumn{3}{c|}{\textbf{SSIM}}                  \\ \cline{3-14} 
                                  &                                 & \textbf{Basic} & \textbf{Resid.} & \textbf{Dense} & \textbf{Basic} & \textbf{Resid} & \textbf{Dense}             & \textbf{Basic} & \textbf{Resid.} & \textbf{Dense} & \textbf{Basic} & \textbf{Resid.} & \textbf{Dense} \\ \hline
\multirow{6}{*}{\textbf{Div2K}}   & \textbf{1}                      & 0.95          & 0.99             & 1.00          & 0.91          & 0.99             & \multicolumn{1}{c|}{0.99} & 24.52         & 41.68            & 41.60         & 0.70            & 0.96             & 0.95          \\
                                  & \textbf{2}                      & 0.91          & 0.98             & 0.99          & 1.65          & 1.92             & \multicolumn{1}{c|}{1.96} & 24.62         & 38.25            & 39.62         & 0.67          & 0.90             & 0.92          \\
                                  & \textbf{3}                      & 0.82           & 0.92              & 0.94          & 1.92           & 2.52             & \multicolumn{1}{c|}{2.63} & 25.03         & 36.67            & 36.52         & 0.69          & 0.85             & 0.85          \\
                                  & \textbf{4}                      & 0.75          & 0.82             & 0.82          & 1.98          & 2.52             & \multicolumn{1}{c|}{2.53}  & 24.45         & 37.86             & 37.49         & 0.69          & 0.88             & 0.88          \\
                                  & \textbf{5}                      & 0.69          & 0.74             & 0.75           & 1.86          & 2.39             & \multicolumn{1}{c|}{2.50}   & 24.90         & 39.45             & 38.65         & 0.70          & 0.90             & 0.90          \\
                                  & \textbf{6}                      & 0.67           & 0.69             & 0.70          & 2.04          & 2.32             & \multicolumn{1}{c|}{2.44}  & 24.72         & 39.53            & 38.94         & 0.70          & 0.91             & 0.90            \\ \hline
\multirow{6}{*}{\textbf{COCO}}  & \textbf{1}                      & 0.98          & 0.99             & 0.99          & 0.96          & 0.99              & \multicolumn{1}{c|}{0.99} & 31.21         & 41.71            & 42.09          & 0.87          & 0.98             & 0.98          \\
                                  & \textbf{2}                      & 0.97          & 0.99             & 0.99          & 1.88          & 1.97             & \multicolumn{1}{c|}{1.97} & 31.56         & 39.00            & 39.08         & 0.86          & 0.96             & 0.95          \\
                                  & \textbf{3}                      & 0.94          & 0.97             & 0.98          & 2.67          & 2.85             & \multicolumn{1}{c|}{2.87} & 30.16         & 37.38            & 36.93         & 0.83          & 0.93             & 0.92          \\
                                  & \textbf{4}                      & 0.87          & 0.95              & 0.95          & 2.99          & 3.60             & \multicolumn{1}{c|}{3.61}  & 31.12         & 36.98            & 36.94         & 0.83          & 0.92             & 0.92          \\
                                  & \textbf{5}                      & 0.84          & 0.90             & 0.92          & 3.43          & 3.99             & \multicolumn{1}{c|}{4.24}  & 29.73         & 36.69            & 36.61         & 0.80          & 0.90             & 0.91          \\
                                  & \textbf{6}                      & 0.78          & 0.84             & 0.87          & 3.34          & 4.07             & \multicolumn{1}{c|}{4.40} & 31.42         & 36.75            & 36.33         & 0.84          & 0.89             & 0.88          \\ \hline
\end{tabular}
\caption{The relative payload and image quality metrics for each dataset and model variant. The Dense model variant offers the best performance across all metrics in almost all experiments.\vspace{1.25em}}
\label{table:metrics}
\end{table*}

We immediately observe that all variants of our model perform better on the COCO dataset than the Div2K dataset. This can be attributed to differences in the type of content photographed in the two datasets. Images from the Div2K dataset tend to contain open scenery, while images from the COCO dataset tend to be more cluttered and contain multiple objects, providing more surfaces and textures for our model to successfully embed data.

\begin{figure}[tb!]
\begin{center}
\includegraphics[width=0.32\linewidth]{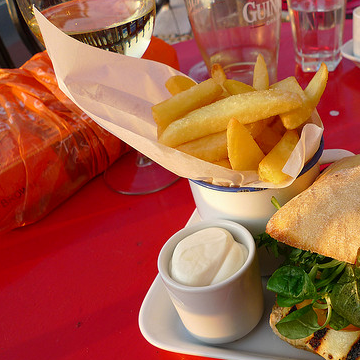}
\includegraphics[width=0.32\linewidth]{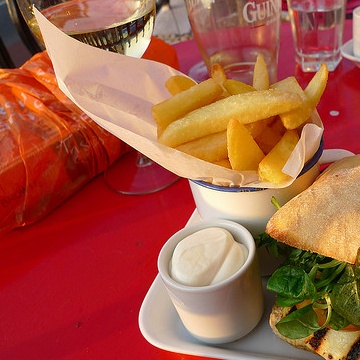}
\includegraphics[width=0.32\linewidth]{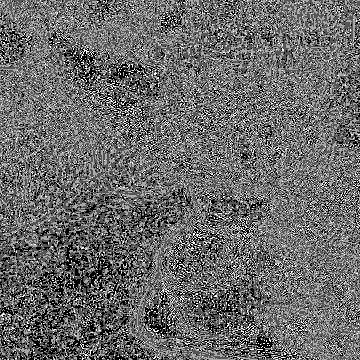}
\includegraphics[width=0.32\linewidth]{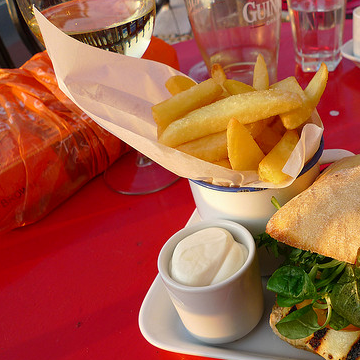}
\includegraphics[width=0.32\linewidth]{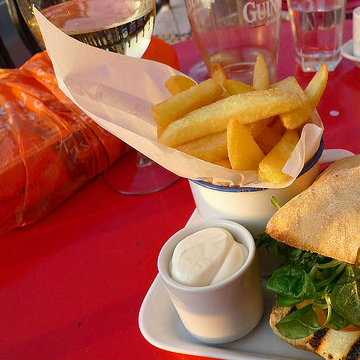}
\includegraphics[width=0.32\linewidth]{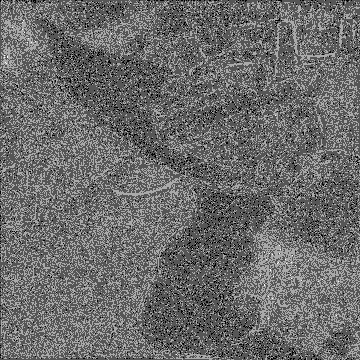}
\caption{A randomly selected pair of cover (left) and steganographic (right) images and the differences between them. The top row shows the output from a simple least-significant-bit steganography algorithm \cite{survey} while the bottom row shows the output from \steg with \bestbpp \bpp. Note that \steg is able to adapt to the image content.}
\label{fig:algorithm_diffs}
\end{center}
\end{figure}

In addition, we note that our dense variant shows the best performance on both relative payload and image quality, followed closely by the residual variant which shows comparable image quality but a lower relative payload. The basic variant offers the worst performance across all metrics, achieving relative payloads and image quality scores that are 15-25\% lower than the dense variant.

Finally, we remark that despite the increased relative payload, the image similarity as measured by the average peak signal to noise ratio between the cover image and the steganographic images produced by the Dense models are comparable to that presented in \cite{Zhu}.

\section{Detecting Steganographic Images}\label{detection}
Steganography techniques are also typically evaluated by their ability to evade detection by steganalysis tools. In this section, we experiment with two open source steganalysis algorithms and measure our model's ability to generate undetectable steganographic images.

\subsection{Statistical Steganalysis}
We use a popular open-source steganalysis tool called StegExpose \cite{StegExpose} which combines several existing steganalysis techniques including Sample Pairs \cite{samplepairs}, RS Analysis \cite{rsanalysis}, Chi Squared Attack \cite{chisquare}, and Primary Sets \cite{primarysets}. To measure the effectiveness of our method at evading detection by these techniques, we randomly select 1,000 cover images from the test set, generating the corresponding steganographic images using our Dense architecture with data depth $6$, and examine the results using StegExpose.

\begin{figure}[htb!]
\begin{center}
\begin{tikzpicture}[scale=0.9]
\begin{axis}[
xlabel={False Positive Rate},
ylabel={True Positive Rate},
xmin=0.0,xmax=1.0,
ymin=0.0,ymax=1.0,
ymajorgrids=true,grid style=dashed,
tick align = outside,xtick pos=left,ytick pos=left
]
\addplot[color=blue,solid] table[x=fpr,y=tpr,col sep=comma]{images/auroc.csv};
\draw[domain=0:1,smooth,variable=\x,gray]  plot ({\x},{\x});
\end{axis}
\end{tikzpicture}
\caption{The receiver operating characteristic (ROC) curve produced by the StegExpose library for a set of 1000 steganographic images generated using the Dense architecture with a data depth of 6. The StegExpose library includes multiple steganalysis tools including $\mathtt{Sample Pairs}$ \cite{samplepairs}, $\mathtt{RS Analysis}$ \cite{rsanalysis}, $\mathtt{Chi Squared Attack}$ \cite{chisquare}, and $\mathtt{Primary Sets}$ \cite{primarysets}. The tool achieves an auROC of 0.59.} 
\label{fig:auroc}
\end{center}
\end{figure}
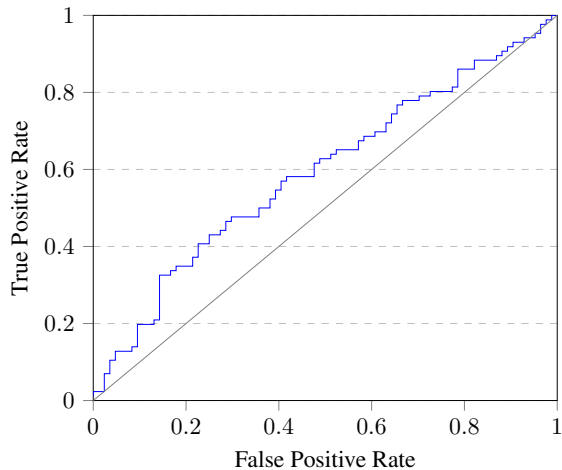

The receiver operating characteristic curve for our Dense model is shown in Figure~\ref{fig:auroc} and we note that the StegExpose tool is only slightly more effective than random guessing with an area under the ROC curve of $0.59$, even for payloads of up to \bestbpp bits per pixel. This shows that our model can successfully evade standard steganalysis tools, meeting the minimum requirement for being a viable steganography algorithm.

\subsection{Neural Steganalysis} 

Recent studies have shown promising results in detecting steganographic images using deep learning based approaches \cite{deep_steganalysis}. Therefore, we proceed to examine whether our model can evade deep learning-based steganalysis tools. We use the model proposed by Ye et al., 2017 in \cite{deep_steganalysis} for steganalysis, with a slight modification to enable support of color images, and train it to detect steganographic images generated by \steg.

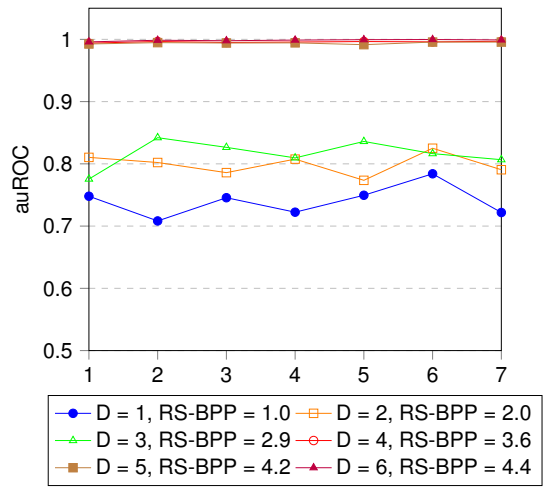
\begin{figure}[!htb]
\begin{center}
\begin{tikzpicture}[scale=0.8,font=\sffamily\sansmath]
\begin{axis}[
xlabel={Number of Instances},
ylabel={auROC},
xmin=1,xmax=7,
ymin=0.5,ymax=1.05,
ymajorgrids=true,grid style=dashed,
tick align = outside,xtick pos=left,ytick pos=left,
legend style={at={(0.5,-0.13)},anchor=north},legend columns=2,legend cell align=left,
]
\addplot[color=blue,mark=*,solid] table[x=num_samples,y=1, col sep=comma] {images/auroc_bpp.csv};
\addplot[color=orange,mark=square,solid] table[x=num_samples,y=2, col sep=comma] {images/auroc_bpp.csv};
\addplot[color=green,mark=triangle,solid] table[x=num_samples,y=3, col sep=comma] {images/auroc_bpp.csv};
\addplot[color=red,mark=o,solid] table[x=num_samples,y=4, col sep=comma] {images/auroc_bpp.csv};
\addplot[color=brown,mark=square*,solid] table[x=num_samples,y=5, col sep=comma] {images/auroc_bpp.csv};
\addplot[color=purple,mark=triangle*,solid] table[x=num_samples,y=6, col sep=comma] {images/auroc_bpp.csv};
 \legend{{D = 1, RS-BPP = 1.0}, {D = 2, RS-BPP = 2.0}, {D = 3, RS-BPP = 2.9}, {D = 4, RS-BPP = 3.6}, {D = 5, RS-BPP = 4.2}, {D = 6, RS-BPP = 4.4}}
\end{axis}
\end{tikzpicture}
\caption{This plot shows the performance of the steganography detector on a held-out test set. The x-axis indicates the number of different \steg instances that were used, while the y-axis indicates the area under the ROC curve.
}
\label{fig:auroc_bpp}
\end{center}
\end{figure}

In a typical scenario, the party that is attempting to detect steganographic images has access to the algorithm used to create them - in our case, an instance of \steg which is parameterized by the pretrained model weights but not the exact model. Using the Dense architecture from Section~\ref{encoder} and the COCO dataset, we attempt to replicate this scenario with the following experimental setup:

\begin{enumerate}
    \item We train $N$ instances of the Dense \steg architecture with different random seeds.
    \item For each of these trained models, we generate a set of 1,000 steganographic images.
    \item Holding out the images generated by the $N$th model as a test set, we train the steganalysis model proposed in \cite{deep_steganalysis} on increasing subsets of the remaining images $\{1\}, \{1, 2\}, \{1, 2, 3\}, ..., \{1, 2, 3, ..., N-1\}$.
    \item We repeat each experiment 3 times and report the average area under the receiver operating curve in Figure~\ref{fig:auroc_bpp}.
\end{enumerate}

This emulates a realistic setting - the party creating the automatic detection model will not have access to the specific \steg model in use, but may have access to the software used to train the models. Therefore, we pose the following question: If the external party does not know the specific model weights but does know the algorithm for generating models, can they detect steganographic images generated by \steg?

Figure \ref{fig:auroc_bpp} shows the performance of our detector for various relative payloads and training set sizes. First, we note that the detector performance, as measured by the area under the receiver operating characteristic (auROC), increases as we increase the number of bits-per-pixel encoded in the image. In addition, we highlight the fact there is no clear trend in the area under the ROC curve as we increase the number of \steg models used for training. This suggests that the external party will have a difficult time building a model which can detect steganographic images generated by \steg without knowing the exact model parameters.

Finally, we compare the detection error for images generated by \steg against those reported by \cite{deep_steganalysis} on images generated by three state-of-the-art steganography algorithms: WOW \cite{wow}, S-UNIWARD \cite{suniward}, and HILL \cite{hill}. Note that these techniques are evaluated on different dataset and as such, the results are only approximate estimates of the actual relative payload achievable on a particular dataset. For a fixed detection error rate of 20\%, we find that WOW is able to encode up to 0.3 bpp, S-UNIWARD is able to encode up to 0.4 bpp, HILL is able to encode up to 0.5 bpp, and \steg is able to encode up to 2.0 bpp.

\section{Related Work} \label{related}
In this section, we describe a few traditional approaches to image steganography and then discuss recent approaches developed using deep learning.

\subsection{Traditional Approaches}
A standard algorithm for image steganography is "Highly Undetectable steGO" (HUGO), a cost function-based algorithm which uses handcrafted features to measure the distortion introduced by modifying the pixel value at a particular location in the image. Given a set of $N$ bits to be embedded, HUGO uses the distortion function to identify the top $N$ pixels that can be modified while minimizing the total distortion across the image \cite{Pevny}.

Another approach is the JSteg algorithm, which is designed specifically for JPEG images. JPEG compression works by transforming the image into the frequency domain using the discrete cosine transform and removing high-frequency components, resulting in a smaller image file size. JSteg uses the same transformation into the frequency domain, but modifies the least significant bits of the frequency coefficients \cite{Li}.

\subsection{Deep Learning for Steganography}
Deep learning for image steganography has recently been explored in several studies, all showing promising results. These existing proposals range from training neural networks to integrate with and improve upon traditional steganography techniques \cite{Tang} to complete end-to-end convolutional neural networks which use adversarial training to generate convincing steganographic images \cite{Hayes,Zhu}.

\noindent \textbf{Hiding images vs. arbitrary data}: 
The first set of deep learning approaches to steganography were \cite{Baluja, Wu}. Both \cite{Baluja} and \cite{Wu} focus solely on taking a \textit{secret image} and embedding it into a \textit{cover image}. Because this task is fundamentally different from that of embedding arbitrary data, it is difficult to compare these results to those achieved by traditional steganography algorithms in terms of the relative payload. 

Natural images such as those used in \cite{Baluja} and \cite{Wu} exhibit strong spatial correlations, and convolutional neural networks trained to hide images in images would take advantage of this property. Therefore, a model that is trained in such a manner cannot be applied to arbitrary data.

\noindent \textbf{Adversarial training}: The next set of approaches for image steganography are \cite{Hayes, Zhu} which make use of adversarial training techniques. The key differences between these approaches and our approach are the loss functions used to train the model, the architecture of the model, and how data is presented to the network.

The method proposed by \cite{Hayes} can only operate on images of a fixed size. Their approach involves flattening the image into a vector, concatenating the data vector to the image vector, and applying feedfoward, reshaping, and convolutional layers. They use the mean squared error for the encoder, the cross entropy loss for the discriminator, and the mean squared error for the decoder. They report that image quality suffers greatly when attempting to increase the number of bits beyond 0.4 bits per pixel.

The method proposed by \cite{Zhu} uses the same loss functions as \cite{Hayes} but makes changes to the model architecture. Specifically, they ``\textit{replicate the message spatially, and concatenate this message volume to the encoder’s intermediary representation}.'' For example, in order to hide $k$ bits in an $N \times N$ image, they would create a tensor of shape $(k, N, N)$ where the data vector is replicated at each spatial location.

This design allows \cite{Zhu} to handle arbitrary sized images but cannot effectively scale to higher relative payloads. For example, to achieve a relative payload of $1$ bit per pixel in a typical image of size $360 \times 480$, they would need to manipulate a data tensor of size $(172800, 360, 480)$. Therefore, due to the excessive memory requirements, this model architecture cannot effectively scale to handle large relative payloads.

\section{Conclusion}
In this paper, we introduced a flexible new approach to image steganography which supports different-sized cover images and arbitrary binary data. Furthermore, we proposed a new metric for evaluating the performance of deep-learning based steganographic systems so that they can be directly compared against traditional steganography algorithms. We experiment with three variants of the \steg architecture and demonstrate that our model achieves higher relative payloads than existing approaches while still evading detection. 

\section*{Acknowledgements}
The authors would like to thank Plamen Valentinov Kolev and Carles Sala for their help with software support and developer operations and for the helpful discussions and feedback. Finally, the authors would like to thank Accenture for their generous support and funding which made this research possible.

\bibliography{references}
\bibliographystyle{arxiv_icml}

\end{document}